\documentclass[]{jingdong}

\usepackage[utf8]{inputenc}
\usepackage[T1]{fontenc}

\usepackage{amsfonts}
\usepackage{amsmath}
\usepackage{amssymb}
\usepackage{algorithm}
\usepackage{algpseudocode}
\usepackage{booktabs}
\usepackage{xcolor}
\usepackage{enumitem}
\usepackage{inconsolata}
\usepackage{listings}
\usepackage{nicefrac}
\usepackage{siunitx}
\usepackage{url}
\usepackage{xspace}

\crefformat{section}{\S#2#1#3}
\Crefformat{section}{\S#2#1#3}
\crefmultiformat{section}{\S#2#1#3}{ and \S#2#1#3}{, \S#2#1#3}{ and \S#2#1#3}
\Crefmultiformat{section}{\S#2#1#3}{ and \S#2#1#3}{, \S#2#1#3}{ and \S#2#1#3}
\crefrangeformat{section}{\S#3#1#4 to \S#5#2#6}
\Crefrangeformat{section}{\S#3#1#4 to \S#5#2#6}

\definecolor{templatekeyword}{HTML}{E1251B}
\definecolor{templatestring}{HTML}{0B7A75}
\definecolor{templatecomment}{HTML}{6B7280}
\definecolor{templatecodebg}{HTML}{F7F7F8}

\lstdefinestyle{templatecode}{
  basicstyle=\ttfamily\small,
  columns=fullflexible,
  backgroundcolor=\color{templatecodebg},
  frame=single,
  rulecolor=\color{black!15},
  numberstyle=\tiny\color{gray},
  keywordstyle=\color{templatekeyword},
  commentstyle=\color{templatecomment},
  stringstyle=\color{templatestring},
  showstringspaces=false,
  tabsize=2,
  breaklines=true,
  breakatwhitespace=true,
  captionpos=b,
  xleftmargin=3.4pt,
  xrightmargin=3.4pt
}

\title{EasyVideoR1: Easier RL for Video Understanding}
\author[1,2,*]{Chuanyu Qin}
\author[1,2,*]{Chenxu Yang}
\author[3,*,\ddagger]{Qingyi Si}
\author[1,2,*]{Naibin Gu}
\author[1,2,*]{Dingyu Yao}
\author[1,2,\dagger]{Zheng Lin}
\author[1,2]{Peng Fu}
\author[3]{Nan Duan}
\author[3]{Jiaqi Wang}

\affiliation[1]{Institute of Information Engineering, Chinese Academy of Sciences, Beijing, China}
\affiliation[2]{School of Cyber Security, University of Chinese Academy of Sciences, Beijing, China}
\affiliation[3]{JD.COM}
\checkdata[Code]{\href{https://github.com/cyuQ1n/EasyVideoR1}{https://github.com/cyuQ1n/EasyVideoR1}}

\contribution[*]{Equal contribution}
\contribution[\dagger]{Corresponding author}
\contribution[\ddagger]{Project lead}
\checkdata[Email]{\email{\{qinchuanyu,yangchenxu,linzheng\}@iie.ac.cn}; \email{siqingyi.phoebus@jd.com}}

\abstract{
Reinforcement learning from verifiable rewards (RLVR) has demonstrated remarkable effectiveness in improving the reasoning capabilities of large language models. As models evolve into natively multimodal architectures, extending RLVR to video understanding becomes increasingly important yet remains largely unexplored, due to the diversity of video task types, the computational overhead of repeatedly decoding and preprocessing high-dimensional visual inputs, and the difficulty of reproducible evaluation across numerous sensitive hyperparameters. Existing open-source RL training frameworks provide solid infrastructure for text and image scenarios but lack systematic optimizations tailored for video modality. In this work, we present \textbf{EasyVideoR1}, a complete and efficient reinforcement learning framework specifically designed for training large vision-language models on video understanding tasks. EasyVideoR1 makes the following contributions: (1) a full video RL training pipeline with offline preprocessing and tensor caching that eliminates redundant video decoding and yields a 1.47 $\times$ throughput improvement; (2) a comprehensive, task-aware reward system covering 11 distinct video and image problem types with unified routing and modular extension; (3) a mixed offline-online data training paradigm that combines curated high-quality trajectories with on-policy exploration, benefiting the learning of more challenging tasks; (4) joint image-video training with independently configurable pixel budgets, allowing the two modalities to mutually reinforce each other; and (5) an asynchronous multi-benchmark evaluation framework covering 22 mainstream video understanding benchmarks, with reproduced accuracy closely aligned with officially reported scores. With 32 H200 GPUs and approximately 20 hours of RL training, Qwen3-VL-8B-Instruct surpasses Qwen3-VL-8B-Thinking on multiple video understanding benchmarks. We welcome valuable contributions from the community.
}

\begin{document}
\maketitle

\section{Introduction}

Reinforcement learning from verifiable rewards (RLVR), exemplified by GRPO \citep{grpo}, has proven highly effective in improving the reasoning capabilities of large language models, as demonstrated by DeepSeek-R1 \citep{deepseekr1}. As the field advances, large language models are rapidly evolving into natively multimodal architectures, such as Qwen3.5 \citep{qwen3.5} and Kimi-K2.5 \citep{kimiteam2026kimik25visualagentic}, which naturally calls for RL training frameworks to go beyond text-only settings and accommodate multimodal scenarios, including modality-aware reward design, efficient visual data processing, and scalable optimization across heterogeneous inputs.

Among the diverse multimodal capabilities, video understanding stands out for its rich real-world applications, spanning embodied intelligence, interactive video dialogue, surveillance analysis, and autonomous driving. Compared to image or text modalities, video introduces unique challenges across all the dimensions mentioned above: reward design must accommodate a broad spectrum of tasks ranging from multiple-choice question answering and optical character recognition to temporal event localization, spatial grounding, object tracking, and dense pixel-level segmentation; visual data processing involves costly preprocessing pipelines such as frame sampling, resizing, and normalization that can become a significant bottleneck; and training must handle the mixing of heterogeneous data sources as well as substantially longer input contexts. While RLVR has been extensively studied for text-based reasoning, applying it systematically to video-language models remains largely unexplored.

Several frameworks have emerged to support RL training for vision-language models. EasyR1 \citep{zheng2025easyr1} provides a clean and scalable foundation extending veRL \citep{sheng2024hybridflow} to support both text and image modalities, with efficient distributed execution via FSDP and vLLM. R1-V \citep{chen2025r1v} demonstrates that RL training on visual counting and geometric reasoning tasks can be achieved at extremely low cost using DeepSpeed. OneThinker \citep{feng2025onethinker} proposes a unified model for vision and video task types to handle heterogeneous reward signals across tasks. However, these frameworks either primarily focus on image-text RL scenarios or lack systematic optimizations tailored for video modality. For instance, both EasyR1 and OneThinker redundantly decode and preprocess video data up to three times across pipeline stages, significantly slowing down training throughput. Moreover, none of them provides evaluation code that faithfully reproduces the accuracy reported by upstream model releases. This is particularly problematic for video benchmarks, where evaluation is sensitive to numerous hyperparameters such as frame sampling strategy, maximum visual token budget, frames per second, the trade-off between frame count and frame rate for long videos, input resolution, and prompt template. Suboptimal choices in any of these can significantly underestimate baseline accuracy, and the sheer volume of visual tokens further makes thorough evaluation slow and costly. There remains a need for a \textbf{complete, efficient framework} specifically designed for the full scope of video understanding, including video-specific preprocessing acceleration, a comprehensive multi-task reward library, joint image-video training, and large-scale asynchronous evaluation with reproducible accuracy.

We present \textbf{EasyVideoR1}, an open-source RL training framework built on EasyR1 that addresses these gaps, with the hope of lowering the barrier and fostering broader community exploration of RL-driven video understanding. EasyVideoR1 makes the following contributions:

\begin{itemize}
\item  \textbf{Complete Video RL Pipeline.} We systematically adapt every stage of the RL training pipeline for video, including metadata-consistent spatial-temporal positional encoding, mixed-modality forward passes under FSDP, independent image/video resolution budgets, and an offline preprocessing cache that yields a 1.47$\times$ throughput improvement on 32 GPUs (\cref{sec:efficiency}).

% \item  \textbf{Comprehensive Task-Aware Reward System.} We provide a unified reward management system that covers 11 distinct task types such as question answering, temporal grounding, spatial localization, and regression. Each task type has a dedicated prompt template, structured output format specification, and answer parser, ensuring accurate reward computation for heterogeneous training data.

\item \textbf{Mixed Offline and Online Data Training.} We extend the EasyR1 training loop to support simultaneous use of pre-collected offline trajectories and online rollout-generated data within a single training iteration. This hybrid regime allows practitioners to leverage curated high-quality datasets while retaining the exploration benefits of on-policy sampling, and is especially beneficial for learning more challenging tasks where on-policy data alone may provide insufficient reward signal.

\item \textbf{Joint Image-Video Training.} We support training batches containing both static images and video clips, with independently configurable pixel budgets for each modality. We handle the engineering challenges of mixed-modality micro-batches under FSDP, ensuring all parameters participate in every forward pass regardless of the modality composition within each micro-batch.

\item \textbf{Asynchronous Multi-Benchmark Evaluation Framework.} We provide an evaluation pipeline based on vLLM's \texttt{AsyncLLMEngine} that supports concurrent inference and result streaming across 22 mainstream video benchmarks (with continuous expansion ongoing), with evaluation protocols adapted to diverse task types. The reproduced accuracy closely aligns with the scores reported in upstream model releases. Taking LVBench as an example, our pipeline achieves approximately $6\sim 7\times$ speedup over vanilla inference frameworks.

\end{itemize}

This report presents the design details of EasyVideoR1 and empirically validates the performance and effectiveness of the framework. With 32 H200 GPUs and approximately 20 hours of RL training, Qwen3-VL-8B-Instruct surpasses Qwen3-VL-8B-Thinking on multiple video understanding benchmarks. EasyVideoR1 is fully open-sourced, and we welcome any valuable contributions from the community.

\section{Related Work}
\subsection{Vision-Language Models for Video Understanding}
The past two years have witnessed rapid progress in video-language pretraining. Qwen2-VL~\citep{wang2024qwen2vlenhancingvisionlanguagemodels} introduced multimodal rotary position embeddings (M-RoPE) and a naive dynamic resolution mechanism to unify image and video understanding within a single architecture. Qwen2.5-VL~\citep{bai2025qwen25vltechnicalreport} further extended this with dynamic FPS sampling along the temporal axis and demonstrated strong performance on long-video comprehension tasks. LLaVA-Video~\cite{zhang2025llavavideovideoinstructiontuning} demonstrates that high-quality synthetic instruction data can effectively drive strong video understanding performance. VideoLLaMA3~\citep{zhang2025videollama3frontiermultimodal} introduces similarity-based video token compression to handle variable-length visual inputs efficiently. More recent work has pushed the frontier further. More recent work has pushed the frontier further. Qwen3-VL~\citep{bai2025qwen3vltechnicalreport} introduces enhanced interleaved-MRoPE for spatial-temporal modeling and explicit textual timestamp tokens replacing T-RoPE for more precise video temporal grounding. Kimi K2.5~\citep{kimiteam2026kimik25visualagentic} extends MoonViT~\citep{kimiteam2025kimivltechnicalreport} to MoonViT-3D with joint text-vision pretraining and reinforcement learning at approximately 15 trillion mixed tokens. Building on the strong video understanding capabilities established by these pretrained backbones, EasyVideoR1 is designed to unlock further gains through RL post-training, enabling models to reason over diverse video tasks via verifiable reward signals.
\subsection{RL Training Frameworks for Vision-Language Models}
A growing ecosystem of open-source frameworks supports RL post-training for large models, with varying degrees of multimodal support. veRL~\citep{sheng2024hybridflow} introduces a hybrid-engine design that co-locates training and inference to maximize GPU utilization; TRL~\citep{vonwerra2020trl} provides accessible RL implementations tightly integrated with the HuggingFace ecosystem; ROLL~\citep{wang2025reinforcement} targets flexible agentic and multi-turn training scenarios. Although all three have incorporated multimodal support, their core optimizations remain centered on training efficiency, usability, or scenario flexibility rather than modality-specific designs for vision-language tasks. A second group of frameworks places stronger emphasis on multimodal RL training. OpenRLHF~\citep{hu2024openrlhf} offers a high-performance RLHF stack whose modular design enables specialized extensions for multimodal RL. ms-SWIFT~\citep{zhao2024swiftascalablelightweightinfrastructure} provides a one-stop training infrastructure that unifies pretraining, SFT, DPO, and GRPO with multimodal support within a single command-line interface. EasyR1~\citep{zheng2025easyr1} builds on veRL to provide a clean, researcher-friendly framework for multimodal RL training, supporting both text and image modalities via FSDP and vLLM rollout. R1-V~\citep{chen2025r1v} demonstrates that R1-style RL training on visual counting and geometric reasoning tasks can be achieved at extremely low cost using DeepSpeed. However, these systems primarily target image-level understanding, with little or no dedicated support for video modality. OneThinker~\citep{feng2025onethinker} takes a step further by extending EasyR1 into a multi-task RL training framework that jointly optimizes over 10 heterogeneous vision task types including video, yet it still lacks a pipeline purpose-built for the full scope of video understanding: accelerated offline video preprocessing, a comprehensive multi-task video reward library, mixed offline-online training, joint image-video batching with independent resolution control, and asynchronous multi-benchmark evaluation. EasyVideoR1 addresses these gaps as a cohesive, immediately deployable framework for the video RL research community.

\section{System Design}

\begin{figure}[t]
\centering
\includegraphics[width=\linewidth]{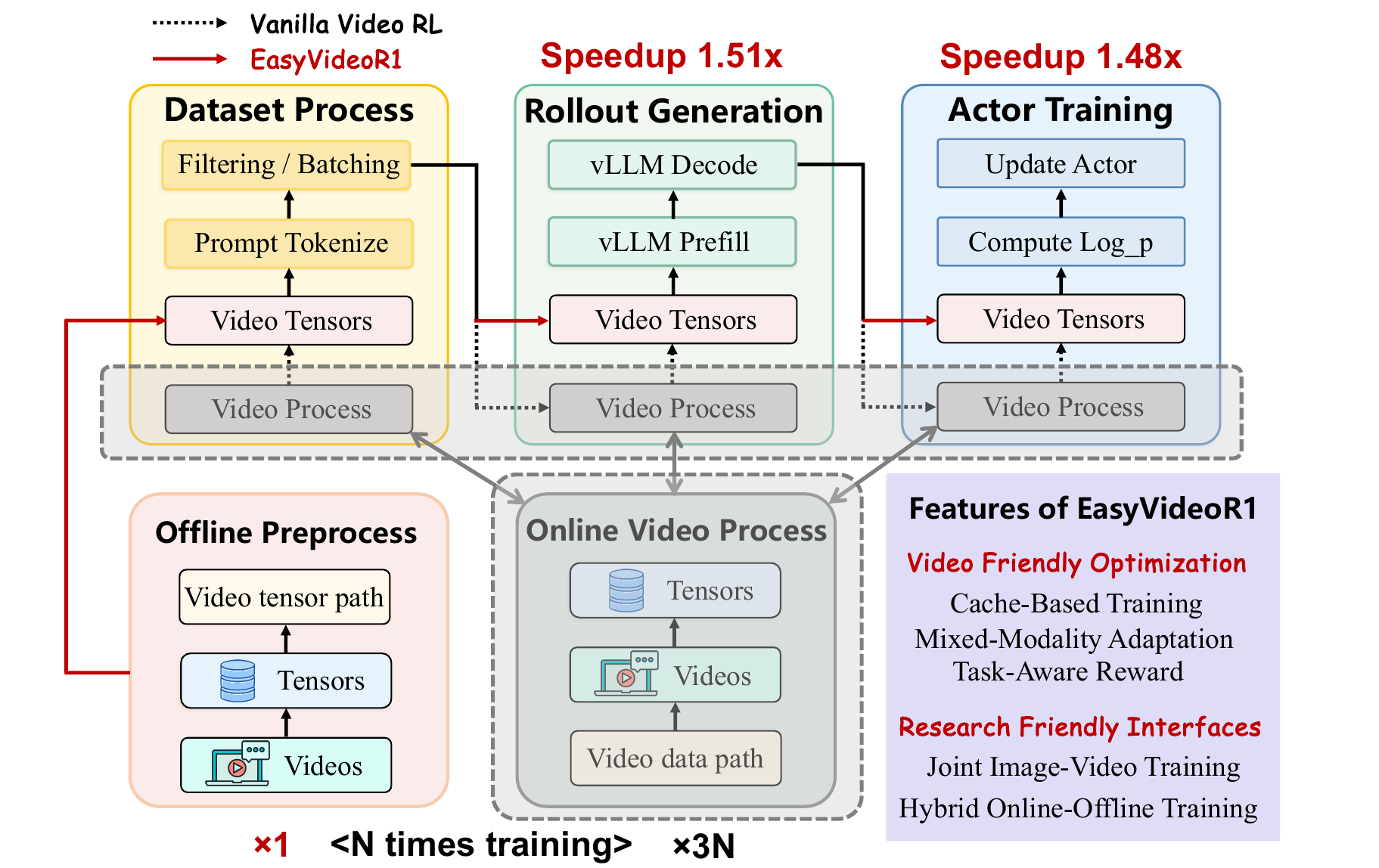}
\caption{Overview of the EasyVideoR1 training pipeline. Videos are preprocessed offline into \texttt{.pt} cache files. During training, each worker loads cached frames locally.}
\label{fig:pipeline}
\end{figure}

% EasyVideoR1 extends EasyR1~\citep{zheng2025easyr1} and veRL~\citep{sheng2024hybridflow} with systematic support for video understanding RL training. We organize our design around three dimensions: adapting the RL pipeline for the video modality (\cref{sec:video_friendly}), providing flexible training paradigms for research iteration (\cref{sec:research_friendly}), and building a high-throughput evaluation framework (\cref{sec:eval_framework}). Figure~\ref{fig:pipeline} provides an overview of the training pipeline.
EasyVideoR1 extends EasyR1~\citep{zheng2025easyr1} and veRL~\citep{sheng2024hybridflow} with systematic support for video understanding RL training and evaluating. We organize our design around three dimensions: adapting the RL pipeline for the video modality (\cref{sec:video_friendly}), providing research-friendly interfaces for algorithm development (\cref{sec:research_friendly}), and building a high-throughput evaluation framework (\cref{sec:eval_framework}). Figure~\ref{fig:pipeline} provides an overview of the training pipeline.

\subsection{Video Friendly Optimization}
\label{sec:video_friendly}

% While recent RL frameworks such as EasyR1~\citep{zheng2025easyr1} and OneThinker~\citep{feng2025onethinker} have begun to support video modality, extending RL training to video understanding still presents several under-addressed challenges: sequence lengths increase by 10--100$\times$  compared to images, video decoding creates CPU-bound I/O bottlenecks, raw video data is not directly passed between pipeline stages but referenced by file paths, requiring redundant decoding and preprocessing at each stage, and the diversity of video tasks demands specialized reward designs. We address these through pipeline-level adaptations (\cref{sec:pipeline_adapt}) and offline preprocessing (\cref{sec:offline_preprocess}).

While recent RL frameworks such as EasyR1~\citep{zheng2025easyr1} and OneThinker~\citep{feng2025onethinker} have begun to support video modality, extending RL training to video understanding still presents several under-addressed challenges: sequence lengths increase by 10--100$\times$ compared to images, video decoding creates CPU-bound I/O bottlenecks, raw video data is not directly passed between pipeline stages but referenced by file paths---requiring redundant decoding and preprocessing at each stage, and mixed image-video scenarios demand modality-aware pipeline adaptations. We address these through offline preprocessing with metadata-consistent caching (\cref{sec:offline_preprocess}), pipeline-level adaptations for mixed-modality training (\cref{sec:pipeline_adapt}), and a task-aware reward system (\cref{sec:reward_system}).

\subsubsection{Efficient RL with Video Caching}
\label{sec:offline_preprocess}

In existing frameworks, each pipeline stage independently decodes raw video files at every training step, making CPU-bound video decoding the dominant throughput bottleneck. EasyVideoR1 decouples video preprocessing from the training loop by providing an offline batch tool that decodes, resamples, and resizes videos into cache files, each keyed by \texttt{(video\_path, fps, max\_frames, max\_pixels)} to automatically invalidate stale entries upon parameter changes. The preprocessing stage is parallelized across multi-worker processes with hash-based deduplication.

Compared to highly compressed video formats such as MP4, directly storing video tensors incurs a disk-space overhead of several orders of magnitude, as modern video codecs achieve compression ratios of several hundred to one. However, by caching frames that have already been temporally sampled and spatially resized, we substantially reduce this storage footprint. In practice, the per-video cache size is bounded by the \texttt{max\_frames} budget: for example, a 
10-minute video sampled at 2\,fps with a cap of 256 frames yields a cache file of roughly 360\,MB regardless of the original video length, compared to tens of megabytes for the compressed source. We view this as a favorable trade-off, exchanging inexpensive storage for expensive GPU-hour throughput gains, and note that adopting a more compact serialization (e.g., uint8 pixels) could further reduce the footprint.

During training, the dataset stage records only the cache file path as a lightweight string rather than the frame tensors themselves. The actual \texttt{.pt} loading occurs locally on each worker at the point of use, reducing inter-node data transfer from megabytes-per-sample tensors to short string paths. When no cache is available, the pipeline falls back transparently to on-the-fly decoding. Because cached frames have already undergone sampling and resizing, subsequent pipeline stages must skip these operations to avoid double processing. EasyVideoR1 achieves this by propagating \texttt{VideoMetadata} (frame rate, sampling indices, spatial dimensions) alongside the cached frames throughout the entire pipeline: metadata is attached during dataset loading, passed to vLLM as \texttt{(tensor, VideoMetadata)} tuples during rollout, and forwarded to the HuggingFace processor during actor training, with \texttt{do\_resize=False} and \texttt{do\_sample\_frames=False} ensuring that each stage produces identical \texttt{video\_grid\_thw} values and thus consistent behaviors. We quantify the throughput improvement from this caching mechanism in \cref{sec:experiments}.

\subsubsection{Mixed-Modality Pipeline Adaptation}
\label{sec:pipeline_adapt}

The RL training pipeline consists of three stages, namely dataset loading, rollout generation (vLLM), and actor training (FSDP), each originally designed for text and image inputs. Supporting mixed image-video training requires coordinated modifications across all stages.

\paragraph{Mixed-Modality Forward Pass.}
LVLM typically processes images and videos through separate encoder branches. In mixed image-video training (\cref{sec:joint_training}), micro-batches containing only one modality leave the other branch inactive, causing FSDP gradient synchronization failures. We resolve this by generating zero-valued dummy tensors for the missing modality and connecting their encoder outputs to the computation graph via zero-weighted addition, ensuring all parameters participate in every forward pass without contributing spurious gradients.

\paragraph{Independent Resolution Budgets.}
Images benefit from high spatial resolution, while videos must balance per-frame resolution against frame count. We decouple the resolution configuration into separate \texttt{image\_max\_pixels}, \texttt{video\_max\_pixels}, and \texttt{video\_max\_frames} parameters, allowing independent tuning of each modality's compute budget.

\subsubsection{Task-Aware Reward System}
\label{sec:reward_system}

Video understanding spans diverse task types requiring specialized scoring logic. EasyVideoR1 provides a modular reward library with a unified routing mechanism: a central dispatcher examines each sample's \texttt{problem\_type} and forwards it to the corresponding reward module. Each task type is implemented as an independent module, enabling incremental extension. Table~\ref{tab:reward_types} summarizes the supported categories. Prompt formatting is handled through Jinja2 templates that are dynamically rendered per task type.

\begin{table}[t]
\centering
\caption{Supported task types and their accuracy scoring methods.}
\label{tab:reward_types}
\small
\setlength{\tabcolsep}{4pt}
\begin{tabular}{lll}
\toprule
\textbf{Category} & \textbf{Task Type} & \textbf{Accuracy Scoring} \\
\midrule
Multiple Choice     & multiple choice               & Exact match       \\
Numerical           & numerical, regression         & Numeric comparison   \\
Temporal Grounding  & temporal grounding            & 1D IoU        \\
ST Grounding        & spatial-temporal grounding    & $0.5\!\times\!\text{tIoU} + 0.5\!\times\!\text{mIoU}$ \\
Spatial Grounding   & spatial grounding             & Bounding-box IoU              \\
Open-ended          & open-ended, video QA          & ROUGE score                   \\
Math                & math                          & Symbolic verification         \\
OCR                 & OCR                           & WER / exact match             \\
Boolean             & boolean                       & Exact match                   \\
Code                & SVG, HTML               & Execution / match             \\
Preference          & LLaVA, critic                 & LLM-as-Judge                  \\
\bottomrule
\end{tabular}
\end{table}

\subsection{Research-Friendly Interfaces for Algorithm Development}
\label{sec:research_friendly}

We identify two research directions with significant potential for advancing video RL: (1) hybrid training that combines offline trajectory data with online rollouts to improve sample efficiency and mitigate cold-start issues, and (2) joint image-video training that leverages abundant image data to strengthen visual reasoning while learning video-specific temporal understanding. To lower the barrier for the community to explore these directions, EasyVideoR1 provides lightweight, ready-to-use interfaces for both paradigms.

\subsubsection{Hybrid Online-Offline Training}
\label{sec:mix_policy}

Standard on-policy methods such as GRPO require all trajectories to be generated by the current policy. In practice, this purely online regime suffers from a cold-start problem, where early rollouts yield sparse reward signals and inefficient gradient estimates, and wastes high-quality trajectory data that may already be available from stronger models or prior checkpoints. Recent work addresses this through various hybrid paradigms: off-policy guidance~\citep{luffy2025}, prefix-based blending~\citep{prefixrft2025,trapo2025}, experience replay~\citep{exgrpo2025,rlep2025}, and adaptive SFT interleaving~\citep{relift2025}.

EasyVideoR1 implements a lightweight \emph{mix-policy} interface: each training sample may carry a pre-collected offline trajectory. During rollout, the framework generates $n{-}1$ on-policy responses and substitutes the final slot with the offline trajectory, assembling a group of $n$ responses that proceeds through reward computation and GRPO update as usual. The mechanism is controlled by a single flag (\texttt{enable\_mix\_policy}) and an optional quality threshold. It operates entirely at the rollout layer without modifying the GRPO algorithm, accepts offline trajectories from any source, and recovers standard on-policy training when disabled.

%\subsubsection{Joint Image-Video Training}
%\label{sec:joint_training}

%High-quality annotated video data remains scarce relative to image QA datasets. Inspired by OneThinker\citep{feng2025onethinker}, joint training leverages abundant image data to strengthen foundational visual reasoning while simultaneously learning video-specific temporal understanding, allowing the two modalities to mutually reinforce each other. Each sample carries a \texttt{data\_type} field that routes it to the appropriate preprocessor and resolution budget. The mixed-modality pipeline adaptations described in \cref{sec:pipeline_adapt}, including independent resolution budgets and DeepStack-aware dummy tensors, provide the necessary engineering foundation for this training paradigm.

\subsubsection{Joint Image-Video Training}
\label{sec:joint_training}
High-quality annotated video data remains scarce relative to image QA datasets. Inspired by OneThinker~\citep{feng2025onethinker}, joint training leverages abundant image data to strengthen foundational visual reasoning while simultaneously learning video-specific temporal understanding, allowing the two modalities to mutually reinforce each other. Each sample carries a \texttt{data\_type} field that routes it to the appropriate preprocessor and decoupled resolution budget, so that image-level and video-level hyperparameters can be tuned independently without interfering with each other. On top of this, we unify the multimodal field schema across image and video samples and keep it consistent throughout data loading, rollout, reward computation, and policy update, which removes modality-conditional branching inside the trainer and makes mixed batches straightforward to assemble. Building on the pipeline adaptations described in \cref{sec:pipeline_adapt}, the current implementation further adopts a strict-failure policy: whenever the number of image or video placeholder tokens does not match the number of visual features produced by the vision encoder, the trainer raises an exception rather than silently truncating or padding. This enforces semantic consistency between textual placeholders and visual features during training, at the cost of requiring upstream data to be well-formed, but in practice surfaces subtle data-processing bugs early and prevents them from corrupting gradient signals over long training runs.

\subsubsection{Broad Model and Algorithm Coverage}
\label{sec:model_algo_coverage}
Beyond the two paradigms above, EasyVideoR1 aims to serve as a general-purpose platform for video RL research by supporting a wide range of vision-language backbones and RL algorithms out of the box.

\paragraph{Advanced VLMs.} The framework natively supports the Qwen2-VL, Qwen2.5-VL, Qwen3-VL, and Qwen3.5 series of vision-language models. Among these, integration for the Qwen3.5 series is contributed by this work, while the remaining backbones are inherited from EasyR1~\citep{zheng2025easyr1}. This coverage spans the most widely used open-source VLM families for video understanding, allowing researchers to directly compare algorithms across different model scales and generations without additional engineering effort.

\paragraph{Rich RL algorithms.} EasyVideoR1 inherits a comprehensive suite of RL algorithms from EasyR1~\citep{zheng2025easyr1}, including GRPO, DAPO, GSPO, CISPO, Reinforce++, ReMax, and RLOO, and additionally contributes new implementations of GDPO and LUFFY \citep{grpo,yu2025dapo,zheng2025groupsequencepolicyoptimization,yang2025dynamicearlyexitreasoning,dai2025sgrpoearlyexitreinforcement,yang2025testtimepromptintervention,yang2026system,yang-etal-2025-weights,yang2026selfdistilledrlvr,liu2026gdpogrouprewarddecouplednormalization,luffy2025}. All algorithms share a unified rollout and reward-computation interface, so switching between them requires only a configuration change. This makes EasyVideoR1 a convenient testbed for studying how different policy-optimization objectives interact with video-specific challenges.

\subsection{Fast \& Comprehensive Evaluation Framework}
\label{sec:eval_framework}

Evaluating multimodal models on video benchmarks at scale poses a fundamental systems challenge: video preprocessing is CPU-intensive and slow, while model inference demands sustained GPU utilization. In naive evaluation pipelines, these two stages execute in strict sequence, resulting in significant hardware underutilization. We design a high-throughput evaluation framework built on top of vLLM's \texttt{AsyncLLMEngine} \citep{kwon2023efficientmemorymanagementlarge}, with the central goal of eliminating CPU--GPU serialization and keeping the GPU saturated throughout the entire evaluation process.

\subsubsection{Asynchronous Inference Design}

Our evaluation framework introduces two key optimizations that collectively transform the evaluation pipeline from a synchronous, batch-oriented workflow into a fully asynchronous streaming architecture.

\paragraph{Precomputed Frame Caching.}
Video preprocessing, including decoding, temporal sampling, and spatial resizing, constitutes the dominant CPU bottleneck in video evaluation. Crucially, these operations are entirely independent of the model and require no GPU involvement, making redundant recomputation across evaluation runs unnecessary. We address this by precomputing all video frames and persisting the results as cache files on disk. At evaluation time, the pipeline reads directly from these cache files, replacing CPU-intensive preprocessing with lightweight file loading and reducing per-video latency from tens of seconds to the millisecond level. Each cache entry is keyed by a tuple of the video path, target frame count, sampling fps, and spatial resolution, so any change in preprocessing parameters automatically invalidates the corresponding cache.  To accelerate the initial cache construction over a large video corpus, we further parallelize the preprocessing stage by dispatching videos to $N$ independent worker processes, each performing decoding and frame extraction without inter-process synchronization, yielding near-linear speedup with respect to the number of workers.

\paragraph{Asynchronous Pipeline with AsyncLLMEngine.}
While precomputed caching eliminates the cost of repeated video preprocessing, a sequential execution pattern between data loading and model inference still leaves the GPU idle at batch boundaries. The standard synchronous inference interface in vLLM blocks until an entire batch of inputs has been loaded before submitting them for inference, and waits for all outputs before accepting new inputs.
We replace this with a three-stage asynchronous pipeline built around vLLM's asynchronous engine, where each stage operates concurrently. In the IO stage, a background thread pool continuously reads cached video frames from disk and submits the prepared inputs to the inference engine as soon as they are ready, without blocking the main inference loop. In the Prefill stage, the engine immediately begins processing each newly arrived input sequence and constructing its key-value cache, without waiting for other in-flight requests to complete. In the Decode stage, requests that have finished prefill immediately enter autoregressive token generation, and their decode steps interleave with the prefill computation of subsequently arrived requests within the same GPU scheduling step. The three stages operate in a fully overlapped fashion: at any given moment, the IO stage may be loading the $(N+2)$-th sample, the Prefill stage processing the $(N+1)$-th, and the Decode stage generating tokens for the $N$-th, keeping the GPU productive at every scheduling step. To further prevent long video sequences from monopolizing the GPU during prefill, we enable chunked prefill, which partitions each long input into fixed-size token chunks so that the scheduler can pack prefill chunks and decode tokens together in every step, maintaining consistently high GPU occupancy regardless of input sequence length.

Together, these two mechanisms ensure that the GPU remains productive at every scheduling step: cached I/O feeds data continuously, asynchronous queuing removes batch-boundary stalls, and chunked prefill prevents any single long sequence from monopolizing compute. Taking LVBench as an example, our pipeline achieves approximately $6\sim 7\times$ speedup over vanilla inference frameworks.

\subsubsection{Supported Benchmarks}

To enable comprehensive and reproducible evaluation, our framework provides a unified interface that currently integrates 22 video understanding benchmarks, spanning six categories: general video understanding, long video understanding, video reasoning, STEM knowledge, spatial understanding, (spatio-)temporal grounding, and streaming video. Table~\ref{tab:benchmarks} summarizes the full list of supported benchmarks along with their task types, scales, and evaluation metrics. These benchmarks collectively cover a wide spectrum of capabilities, from fine-grained motion perception and temporal reasoning to expert-level knowledge question answering and spatio-temporal localization, providing a thorough assessment of model strengths and weaknesses across diverse video understanding dimensions.
Each benchmark is registered through a lightweight configuration that specifies the data loading logic, prompt formatting, answer extraction, and scoring function. Adding a new benchmark requires only implementing a thin adapter that conforms to this interface, without modifying any core pipeline code. This modular design allows researchers to evaluate models across the entire benchmark suite with a single command and a consistent set of inference configurations, facilitating fair and reproducible comparisons. We have verified that the accuracy produced by our framework closely matches the officially reported scores across all supported benchmarks.

\begin{table}[t]
\centering
\caption{Video understanding benchmarks supported by the evaluation framework.}
% Multiple Choice: multiple-choice; Open-ended: open-ended; Regression: regression; Numerical: numerical.}
\label{tab:benchmarks}
\small
\setlength{\tabcolsep}{4pt}
\begin{tabular}{llrr}
\toprule
\textbf{Benchmark} & \textbf{Task Type} & \textbf{Number} & \textbf{Metric} \\
\midrule
\multicolumn{4}{l}{\textit{General Video Understanding}} \\
Video-MME \citep{fu2025videommefirstevercomprehensiveevaluation}        & Multiple Choice    & 2,700 & Accuracy            \\
Video-MME-v2 \citep{fu2026videommev2stagebenchmarkscomprehensive}        & Multiple Choice    & 3,200 & Accuracy            \\
MVBench \citep{li2024mvbenchcomprehensivemultimodalvideo}         & Multiple Choice    & 3,586 & Accuracy            \\
TempCompass \citep{liu2024tempcompassvideollmsreally}     & Multiple Choice    & 7,540 & Accuracy            \\
MotionBench  \citep{hong2025motionbenchbenchmarkingimprovingfinegrained}    & Multiple Choice    & 3,715 & Accuracy            \\
\midrule
\multicolumn{4}{l}{\textit{Long Video Understanding}} \\
LVBench  \citep{wang2025lvbenchextremelongvideo}        & Multiple Choice    & 1,492 & Accuracy            \\
LongVideoBench \citep{wu2024longvideobenchbenchmarklongcontextinterleaved}  & Multiple Choice    & 1,337 & Accuracy            \\
MLVU \citep{zhou2025mlvubenchmarkingmultitasklong}            & Multiple Choice    &   502 & Accuracy            \\
\midrule
\multicolumn{4}{l}{\textit{Video Reasoning}} \\
Video-Holmes \citep{cheng2025videoholmesmllmthinklike}    & Multiple Choice              & 1,837 & Accuracy            \\
MINERVA \citep{nagrani2025minervaevaluatingcomplexvideo}         & Multiple Choice              & 1,431 & Accuracy            \\
VCR-Bench \citep{qi2025vcrbenchcomprehensiveevaluationframework}       & Multiple Choice + Open-ended         & 1,034 & Accuracy / LLM-as-a-judge \\
VideoReasonBench \citep{liu2026videoreasonbenchmllmsperformvisioncentric} & Open-ended              & 1,440 & LLM-as-a-judge        \\
LongVideo-Reason \citep{chen2025scalingrllongvideos} & Multiple Choice              &   851 & Accuracy            \\
\midrule
\multicolumn{4}{l}{\textit{STEM Knowledge}}   \\
MMVU \citep{zhao2025mmvumeasuringexpertlevelmultidiscipline}            & Multiple Choice + Open-ended             &   1,000 & Accuracy            \\
Video-MMMU \citep{hu2025videommmuevaluatingknowledgeacquisition}      & Multiple Choice              &   900 & Accuracy            \\
VideoMathQA \citep{rasheed2025videomathqabenchmarkingmathematicalreasoning}     & Multiple Choice              & 2,100 & Accuracy            \\
\midrule
\multicolumn{4}{l}{\textit{Spatial Understanding}} \\
VSI-Bench \citep{yang2025thinkingspacemultimodallarge_vsibench}       & Multiple Choice + Regression       & 5,130 & Accuracy      \\
\midrule
\multicolumn{4}{l}{\textit{(Spatio-)Temporal Grounding}} \\
Charades-STA \citep{gao2017talltemporalactivitylocalization}    & Regression            & 3,720 & tIoU            \\
STVG \citep{zhang2020doesexistspatiotemporalvideo}    & Regression            & 2,000 & tIoU + mIoU     \\
\midrule
\multicolumn{4}{l}{\textit{Streaming}} \\
OVOBench \citep{li2025ovobenchfarvideollmsrealworld}    & Multiple Choice + Counting            & 3,035 & Accuracy            \\
ODVBench \citep{zeng2025streamforestefficientonlinevideo}    & Multiple Choice            & 7,896 & Accuracy     \\
LiveSports-QA \citep{chen2025livecclearningvideollm}    & Multiple Choice            & 1,174 & Accuracy     \\
\bottomrule
\end{tabular}
\end{table}
 
\section{Experiments}
\label{sec:experiments}

% We validate EasyVideoR1 along two dimensions: (1) the end-to-end RL training pipeline produces consistent benchmark improvements over the base model, and (2) the offline preprocessing mechanism yields substantial training throughput gains.
We design our experiments to answer two questions: (1) Can an \textbf{instruct} model, after RL training with EasyVideoR1, surpass its corresponding \textbf{ithinking} variant? (2) How much training throughput improvement does the offline preprocessing and caching mechanism yield?

\subsection{Experimental Setup}

% \paragraph{Base Model.}
% We use Qwen3-VL-8B-Instruct~\citep{bai2025qwen3vltechnicalreport} as the base model, which employs the DeepStack architecture with interleaved M-RoPE positional encoding and represents one of the strwongest open-source video-language models at this scale.

% \paragraph{Training Data.}
% We curate a training set of approximately 99K video samples from multiple open-source video reasoning datasets, including OneThinker~\citep{feng2025onethinker}, Video-R1~\citep{feng2025video}, and VideoChat-R1~\citep{li2025videochatr1}. To ensure that training samples lie within the model's learning frontier, we apply a pass-rate-based filtering strategy: for each candidate sample, we perform $k{=}8$ rollouts using the base model and retain only samples with partial success ($0 < \text{pass rate} < 1$), removing trivially solved instances.

\paragraph{Base Model and Training Data.}
EasyVideoR1 natively supports the Qwen2.5-VL and Qwen3-VL model families. We select Qwen3-VL-8B-Instruct~\citep{bai2025qwen3vltechnicalreport} as the representative base model for our experiments, as it employs the DeepStack architecture with interleaved M-RoPE positional encoding and represents one of the strongest and most widely adopted open-source video-language models at this scale. We train on approximately 100K video samples assembled from publicly available video RL datasets such as OneThinker~\citep{feng2025onethinker}, Video-R1~\citep{feng2025video}, and VideoChat-R1~\citep{li2025videochatr1}. To ensure that training samples lie within the model's learning frontier, we apply a pass-rate-based filtering strategy: for each candidate sample, we perform $k{=}8$ rollouts using the base model and retain only samples with partial success ($0 < \text{pass rate} < 1$), removing trivially solved instances.

\paragraph{Training Configuration.}
We train with GRPO using the DAPO clipping variant~\citep{yu2025dapo} (asymmetric clip ratios $\epsilon_{\text{low}}{=}0.2$, $\epsilon_{\text{high}}{=}0.28$) with KL penalty disabled. The rollout group size is $n{=}8$ with a global batch size of 256. We use a constant learning rate of $1{\times}10^{-6}$ with AdamW ($\beta_1{=}0.9$, $\beta_2{=}0.999$, weight decay $0.01$). Video inputs are sampled at 2 FPS with a maximum of 128 frames and a per-frame pixel budget of 262,144; image inputs use a separate budget of 1,048,576 pixels. The maximum response length is 4,096 tokens. Training runs on 32 GPUs with FSDP full sharding, gradient checkpointing, padding-free attention, and dynamic batching enabled. Rollout inference uses vLLM with tensor parallelism size 2.

\paragraph{Evaluation.}
We select 10 representative benchmarks from Table~\ref{tab:benchmarks} spanning four categories: general video understanding (Video-MME, MVBench, TempCompass), long video understanding (LVBench, LongVideoBench, MLVU), video reasoning (Video-Holmes), and STEM knowledge (MMVU, Video-MMMU, VideoMathQA). All evaluations use our asynchronous evaluation framework (\cref{sec:eval_framework}) with greedy decoding.

% \subsection{Main Results}
\subsection{Unlocking the Potential of Instruct Models}

\begin{figure}[t]
\centering
\includegraphics[width=\linewidth]{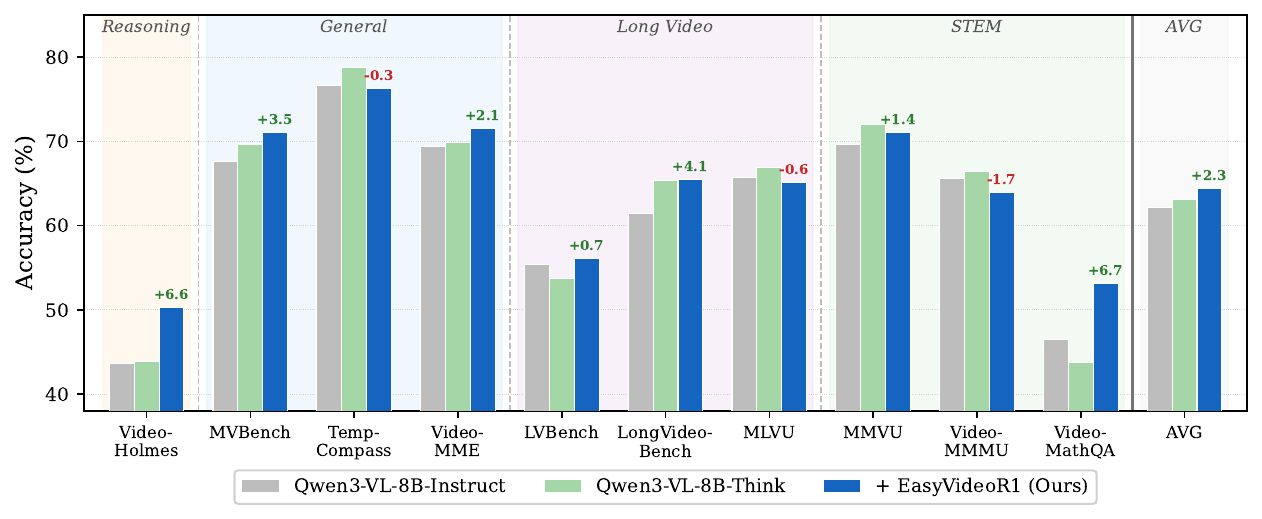}
\caption{Benchmark performance comparison. The number above each blue bar indicates the accuracy change relative to the Instruct baseline. Background colors denote benchmark categories. EasyVideoR1 training yields an average improvement of +2.3 points, with the largest gains on reasoning (+6.6 on Video-Holmes) and mathematical (+6.7 on VideoMathQA) tasks.}
\label{fig:benchmark}
\end{figure}

Figure~\ref{fig:benchmark} compares three model variants across the 10 selected benchmarks: the base Qwen3-VL-8B-Instruct, its officially released thinking variant (Qwen3-VL-8B-Think), and the model after 200 GRPO training steps with EasyVideoR1.

EasyVideoR1 training improves the average accuracy from 62.1 to 64.4 (+2.3), demonstrating that the framework's end-to-end pipeline---from data loading through reward computation to policy update---functions correctly and effectively. Several observations stand out:

\begin{itemize}[nosep,leftmargin=*]
\item \textbf{Reasoning and mathematical tasks benefit most.} Video-Holmes (+6.6) and VideoMathQA (+6.7) show the largest improvements, indicating that RL training effectively strengthens the model's deliberative reasoning capabilities on video inputs.
\item \textbf{General video understanding improves consistently.} Video-MME (+2.1), MVBench (+3.5), and LVBench (+0.7) all show positive gains, confirming that RL training does not degrade broad video comprehension.
\item \textbf{Competitive with the thinking variant.} The RL-trained model achieves comparable or superior results to Qwen3-VL-8B-Think on most benchmarks, while operating in standard (non-thinking) inference mode without additional reasoning overhead.
\end{itemize}

% \subsection{Training Efficiency}
\subsection{How Much Does Offline Preprocessing and Caching Help?}
\label{sec:efficiency}

We compare the training throughput of cache-based loading versus on-the-fly video decoding under identical configurations: Qwen3-VL-8B on 32 GPUs (4 nodes $\times$ 8 GPUs) with a global batch size of 32, video sequences up to 256 frames, and all other hyperparameters held constant. The only variable is the video loading mode: \texttt{prefer\_preprocessed} (cache) versus \texttt{realtime\_only} (on-the-fly). We report metrics averaged over 95 and 68 training steps respectively, excluding the first warmup step.

\begin{figure}[t]
\centering
\includegraphics[width=\linewidth]{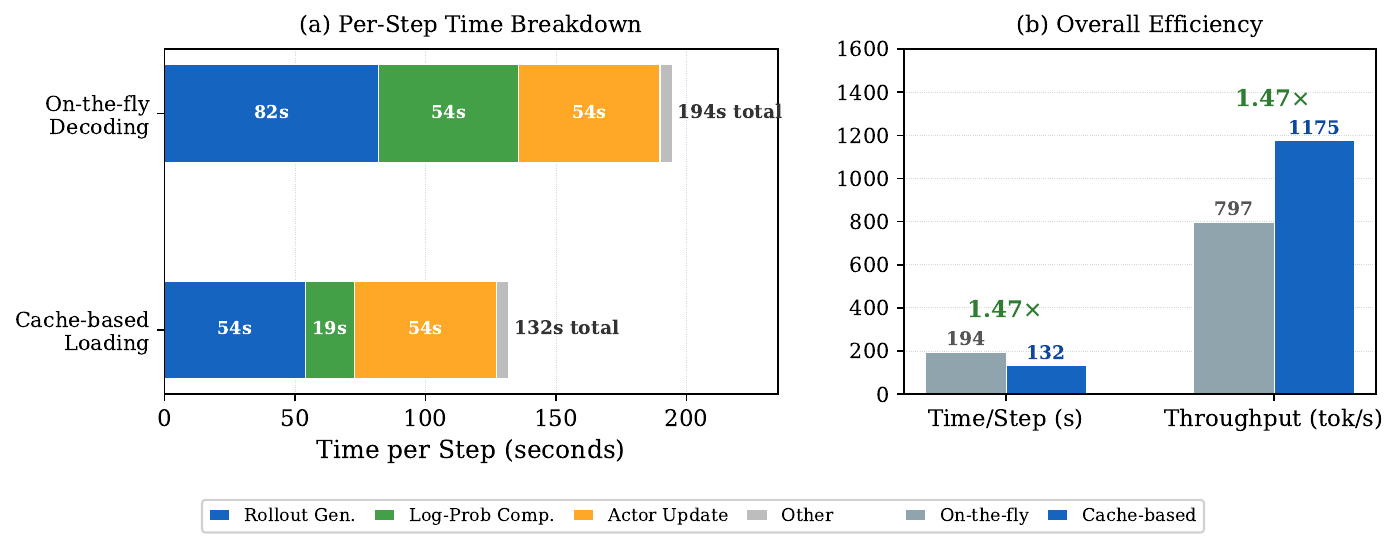}
\caption{Training efficiency comparison between cache-based loading and on-the-fly video decoding. (a) Per-step time breakdown by pipeline phase. Cache-based loading reduces rollout generation time by 1.5$\times$ and reference model forward time by 2.9$\times$, while actor update time remains unchanged. (b) Overall efficiency: cache-based loading achieves a 1.47$\times$ speedup in both wall-clock time per step and token throughput.}
\label{fig:efficiency}
\end{figure}

As shown in Figure~\ref{fig:efficiency}, cache-based loading achieves a \textbf{1.47$\times$ overall speedup}, reducing the average step time from 194.5s to 131.9s and increasing token throughput from 797 to 1{,}175 tokens/s. A phase-level breakdown reveals the source of this improvement:

\begin{itemize}[nosep,leftmargin=*]
\item \textbf{Rollout generation} is reduced from 82.1s to 53.9s \textbf{(1.52$\times$)}, as the vLLM inference engine no longer blocks on CPU-bound video decoding when loading input sequences.
\item \textbf{Reference model forward pass} benefits even more dramatically, dropping from 53.6s to 18.8s \textbf{(2.85$\times$)}. In on-the-fly mode, the reference model stage must independently re-decode the same videos that were already processed during rollout; caching eliminates this redundant computation entirely.
\item \textbf{Actor parameter update} remains constant at ${\sim}$54s in both modes, as expected---this phase operates on token-level gradients and is independent of video I/O.
\end{itemize}

Importantly, the total tokens processed per step are nearly identical (${\sim}$4.93M) under both modes, confirming that the caching mechanism preserves training semantics while delivering acceleration.

\section{Conclusion}

In our pursuit of advancing video understanding through post-training of multimodal LLMs, we found that existing RL frameworks were not particularly well-suited for video understanding scenarios. Therefore, we built EasyVideoR1 upon EasyR1 to implement relevant optimizations, which we have outlined in this report. To the best of our knowledge, this should be the most suitable code repository for research on RL post-training for video understanding at the time of this paper's release. It supports a wide range of video understanding tasks, incorporates research-friendly interfaces (mixed off-policy and on-policy training, joint image-video training), enhances training efficiency for video RL through systematic design, and provides an efficient, comprehensive, and accuracy-aligned evaluation framework. We hope this repository can inspire enthusiasm within the multimodal community for video understanding research. We also call upon community researchers to join us in maintaining this codebase, working together to create the most comprehensive and research-friendly repository for video understanding. We welcome and will consider merging any valuable pull requests.

\section{Acknowledgment}

EasyVideoR1 is built upon several outstanding open-source projects. We sincerely thank the teams behind EasyR1, veRL, and OneThinker for their valuable contributions to the community, which have laid a solid foundation for our work.

\bibliographystyle{assets/plainnat}
\bibliography{references}

%\newpage
%\beginappendix
%\input{sections/99_appendix}

\end{document}